\documentclass{article}

\usepackage[accepted]{icml2025}

\usepackage{microtype}           % Better typography
\usepackage{graphicx}            % Figures
\usepackage{subfigure}           % Subfigures
\usepackage{booktabs}            % Professional tables
\usepackage{hyperref}            % Hyperlinks
\usepackage{amsmath,amssymb}     % Math packages
\usepackage{mathtools}           % Further math tools
\usepackage{amsthm}              % Theorem environments

\usepackage{multirow}            % Multirow in tables
\usepackage{caption}             % Custom captions
\usepackage{cleveref}            % Clever references
\theoremstyle{plain}

\theoremstyle{definition}

\theoremstyle{remark}

\icmltitlerunning{An Iterative Framework for Generative Backmapping of Coarse Grained Proteins}

\begin{document}

\twocolumn[
\icmltitle{An Iterative Framework for Generative Backmapping of Coarse Grained Proteins}

% You may provide any keywords that you find helpful for describing your paper;
% these are used to populate the "keywords" metadata in the PDF but will not 
% be shown in the document.
\icmlkeywords{Machine Learning, Backmapping, Coarse-Grained Modeling, Variational Autoencoders, Proteins}

% NOTE: For double-blind, do NOT include author info here.
% If not double-blind, you can uncomment and fill these lines:

\begin{icmlauthorlist}
\icmlauthor{Georgios Kementzidis}{stonybrookAMS}
\icmlauthor{Erin Wong}{stonybrookGarcia}
\icmlauthor{John Nicholson}{swarthmore}
\icmlauthor{Ruichen Xu}{stonybrookAMS}
\icmlauthor{Yuefan Deng}{stonybrookAMS}
\end{icmlauthorlist}

\icmlaffiliation{stonybrookAMS}{Department of Applied Mathematics and Statistics, Stony Brook University, Stony Brook, NY 11790, USA}
\icmlaffiliation{stonybrookGarcia}{Garcia Center for Polymers at Engineered Interfaces, Stony Brook University, Stony Brook, NY 11790, USA}
\icmlaffiliation{swarthmore}{Swarthmore College, Swarthmore, PA 19081, USA}

\icmlcorrespondingauthor{Yuefan Deng}{yuefan.deng@stonybrook.edu}

\vskip 0.3in
]

% If you use \icmlauthorlist above, then you can do:
\printAffiliationsAndNotice{} 
% or
% \printAffiliationsAndNotice{\icmlEqualContribution} 
% depending on the situation.

\begin{abstract}
The techniques of data-driven backmapping from coarse-grained (CG) to fine-grained (FG) representation often struggle with accuracy, unstable training, and physical realism, especially when applied to complex systems such as proteins. In this work, we introduce a novel iterative framework by using conditional Variational Autoencoders and graph-based neural networks, specifically designed to tackle the challenges associated with such large-scale biomolecules. Our method enables stepwise refinement from CG beads to full atomistic details. We outline the theory of iterative generative backmapping and demonstrate via numerical experiments the advantages of multistep schemes by applying them to proteins of vastly different structures with very coarse representations. This multistep approach not only improves the accuracy of reconstructions but also makes the training process more computationally efficient for proteins with ultra-CG representations.
\end{abstract}

\section{Introduction}
\label{sec: intro}

Coarse-grained (CG) models simplify complex molecular systems by clustering atoms into larger particles, known as CG beads, effectively eliminating certain internal degrees of freedom \cite{1,2}. This reduction leads to fewer interactions, resulting in more efficient computational models that enable spatial and temporal scales inaccessible with fine-grained (FG) simulations.% \cite{3}. 
By internally contracting and consolidating the intricacies of atomic interactions, CG models enable simulations that can provide more meaningful and realistic insights into phenomena at mesoscopic and macroscopic scales \cite{4}. This trade-off, however, comes with the challenge of preserving essential molecular characteristics in the CG representation, such as the structure, dynamics, and energetics of the FG system. Consequently, CG models must be carefully designed to retain these core properties while gaining computational efficiency \cite{5}. Such models are invaluable in many fields including biomolecular research and materials science, where exploring long-timescale processes is critical but computational resources are often limited.

While CG models greatly enhance computational efficiency, significant challenges emerge when reconstructions of the original atomic details are desired. The results of CG simulations often have to be converted back to the FG scale in order for several properties and interatomic interactions to be investigated through atom-level representations. Backmapping (or de-coarsening or inverse coarsening) is a process which seeks to restore atomistic resolution from the CG representation. 
%Achieving accurate backmapping is challenging for three reasons, outlined in previous work \cite{6}. Briefly, first, multiple FG structures can correspond to a single CG structure, making backmapping a stochastic inverse problem. Second, a successful and practical backmapping method must be general enough to perform well for CG mappings with varying resolution. And finally, designing an operator requires respecting the overall geometry of the molecules, as well as equivariance under translations and rotations.

\begin{figure*}[t]
    \centering
    \includegraphics[width=1.0\linewidth]{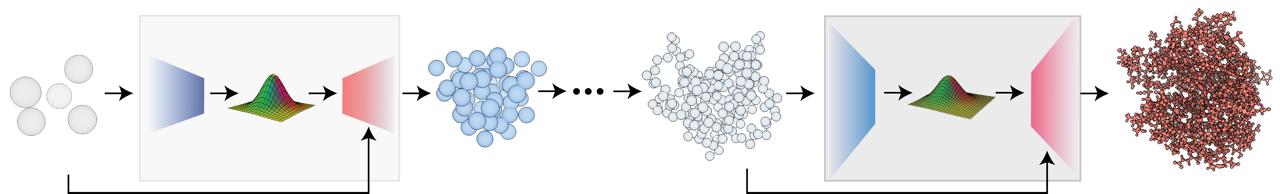}
    % \caption{An illustration of the $k$-step scheme. The blue networks correspond to the independent priors $p_{\psi_i}(\mathbf{z}_i | \mathbf{x}_{i+1})$, while the red networks correspond to the independent decoders $p_{\theta_i}(\mathbf{x}_i | \mathbf{x}_{i+1}, \mathbf{z}_i)$.}
    \caption{An illustration of the multistep generative backmapping scheme. Starting from a UCG structure, we restore atomistic resolution using independent priors (blue networks) and decoders (red networks), one resolution-step at a time.}
    \label{fig:f_3}
\end{figure*}

This problem has been tackled in several ways in the past. One of the earliest successful attempts was given by Wassenaar et. al \cite{7}: initial placement of the atoms followed by force field relaxation restores interatomic distances.% and tertiary structure – in the case of more complex macromolecules like proteins. 
However, this method is very time-consuming because of the need to run all-atom MD. Recently, several ML-based solutions have been proposed. A solution using conditional Generative Adversarial Networks \cite{8,9} was proposed, but is limited to condensed phase systems and uses voxelization. The idea of voxelization also appeared in a model trained to de-coarsen entire trajectories by conditioning the coordinate prediction on the current CG structure and the previous-timestep FG structure \cite{10}. Later advancements using conditional Variational Autoencoders \cite{6,11} introduced equivariant operators with the use of the appropriate graph neural networks (GNN). Lately, the most recent trends include diffusion models \cite{12,13,14}, but they are usually trained and tested on \(C_\alpha\) traces or on CG resolutions like MARTINI \cite{15}, CABS \cite{16}, UNRES \cite{17}, meaning they restore the atomistic details from the residue-based coarse-graining (RBCG) resolution or a higher one – more than one bead per residue. On the other hand, there are models designed to work for arbitrary CG resolutions \cite{6,18}, but they are tested only on either small molecules, or larger molecules with high-resolution CG. Furthermore, previous work emphasizes the chemical transferability of some models \cite{8,11,14}.

Despite the recent advancements and their applicability to a variety of CG resolutions, little attention has been given to ultra-coarse graining (UCG), a regime in which each CG bead corresponds to tens or even hundreds of atoms \cite{19}. The speedup of UCG simulations is more pronounced \cite{20,21,22}, and work on this class of applications, where the current methods underperform, manifests our focuses.

Inspired by super-resolution imaging \cite{23}, we introduce iterative backmapping as a practical tool to alleviate the weaknesses of the previous methods on such CG representations. We develop a theoretical formulation and demonstrate the benefits of our scheme. We verify our hypothesis experimentally by applying a 2-step 
%and 3-step 
generative backmapping scheme to proteins with different structural and functional characteristics. An illustration of the multistep scheme is depicted in Fig.~\ref{fig:f_3}. When compared to the baseline \textsc{CGVAE} model developed by Wang and Bombarelli \cite{6}, the resulting structures are more accurate in restoring the FG structure’s geometry and in conforming with the rules of basic science. 
% Obviously, our multistep method, far from being perfect, requires proper choice of the CG operator, which helps determine the extent to which backmapping, conversely, can restore lost information between two scales.

Our contributions can be summarized as follows:
\begin{itemize}
    \item We identified the weaknesses of existing generative backmapping techniques --- single-step schemes --- on proteins with UCG representation.
    \item We developed a theoretical formulation for iterative generative backmapping of proteins using variational inference, and derive the objective's lower bound, 
    i.e., the loss function of the model.
    % \item We formulated the lower bound of the root mean squared deviation (RMSD) for single-step backmapping between two different scales, a key figure indicating the limitations of any backmapping method.
    \item We demonstrated numerically, through 2-step 
    % and 3-step 
    schemes, the advantages over the state-of-the-art 1-step baselines with a variety of metrics on two very different proteins.
\end{itemize}

\section{Theory}
\label{sec: label}

\begin{figure*}[ht]
    \centering
    \includegraphics[width=0.9\linewidth]{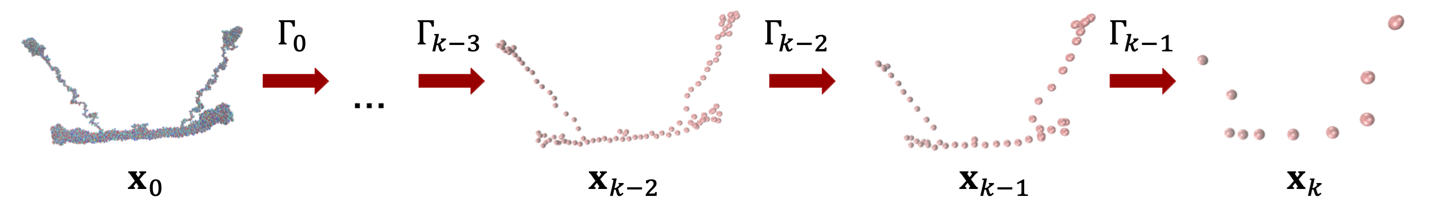} % Replace with your actual figure file
    \caption{An FG conformation $\mathbf{x}_0$ with $k$ progressively coarser and less informative CG 
    conformations $\mathbf{x}_i$. The average CG bead size $\rho$ is increasing.}
    \label{fig:f_1}
\end{figure*}

Consider a molecule with $n_0$ atoms whose coordinates are $\mathbf{x}_0 \in \mathbb{R}^{n_0 \times 3}$. 
Let $k$ be a positive integer, such that $\mathbf{x}_k \in \mathbb{R}^{n_k \times 3}$ is the final CG representation 
with $n_k$ beads, which we aim to reconstruct. As shown in Fig.~\ref{fig:f_1}, 
we consider $k$ coarsening operators $\Gamma_0, \Gamma_1, \ldots, \Gamma_{k-1}$ such that
\[
\Gamma_i(\mathbf{x}_i) = \mathbf{x}_{i+1} \in \mathbb{R}^{n_{i+1} \times 3},
\]
leading to progressively ``coarser'' CG representations $\mathbf{x}_1, \mathbf{x}_2, \ldots, \mathbf{x}_k$ 
with varying CG bead numbers $n_0 > n_1 > n_2 > \cdots > n_k$.

We define the ratio of the number of residues to the number of CG beads (\#~residues / \#~CG~beads) 
as the \emph{average CG bead size} and denote it by $\rho$. Notice that for higher $k$, $\rho$ is higher, 
since each bead represents---on average---a bigger number of residues. 
For reference, $\rho=1$ is equivalent to RBCG, while $\rho \gg 1$ corresponds to some UCG representations.

Given this formulation, $\mathbf{x}_k$ is the result of applying successive coarsening operators to the FG structure:
\[
\mathbf{x}_k 
% = \Gamma_{k-1}\!\bigl(\Gamma_{k-2}(\ldots \Gamma_{1}(\Gamma_{0}(\mathbf{x}_0))\bigr)
= (\Gamma_{k-1} \circ \Gamma_{k-2} \circ \ldots \circ \Gamma_{1} \circ \Gamma_{0})(\mathbf{x}_0).
\]

We define and decompose the conditional distribution $p(\mathbf{x}_0 | \mathbf{x}_k)$ as:
\[
p(\mathbf{x}_0 | \mathbf{x}_k)
% = p_0(\mathbf{x}_0 \mid \mathbf{x}_1)\, p_1(\mathbf{x}_1 \mid \mathbf{x}_2)\, \ldots \, p_{k-1}(\mathbf{x}_{k-1} \mid \mathbf{x}_k)
= \prod_{i=0}^{k-1} p_i(\mathbf{x}_i |\mathbf{x}_{i+1}).
\]
The distribution is factorized into multiple conditional distributions, considering the intermediate CG representations. The Markov property, which dictates conditional independence between a representation $\mathbf{x}_i$ and its non-immediate coarser versions $\mathbf{x}_j, j>i+1$, is required for this factorization to hold. In backmapping, given a representation $\mathbf{x}_i$ 
of some resolution, the property implies that any additional knowledge of non-adjacent coarse 
representations $\mathbf{x}_j, j>i+1$ does not provide more information about $\mathbf{x}_i$ 
than the adjacent coarse representation $\mathbf{x}_{i+1}$. 
For instance, in Fig.~\ref{fig:f_1}, $\mathbf{x}_{k-1}$ is more informative than $\mathbf{x}_k$. 
In principle, this property is not guaranteed, which can be rectified by defining 
appropriate coarsening operators $\Gamma_i$ through topology-conserving algorithms 
to preserve the FG molecule's geometry~\cite{6,24,25,26}.

First, we focus on $p_i(\mathbf{x}_i | \mathbf{x}_{i+1})$, the probability of reconstructing a true, high-resolution $\mathbf{x}_i$ given an input CG $\mathbf{x}_{i+1}$. 
Employing variational inference~\cite{27} and following~\cite{6}, we parametrize the conditional distribution by introducing the approximate posterior distribution 
$q_{\phi_i}(\mathbf{z}_i | \mathbf{x}_i, \mathbf{x}_{i+1})$ with parameters $\phi_i$, 
where $\mathbf{z}_i$ denotes a latent space variable. Also, we describe the original conditional 
distribution as an integral over the latent space with
\[
p_i(\mathbf{x}_i | \mathbf{x}_{i+1}) 
= \int p_{\theta_i}(\mathbf{x}_i | \mathbf{x}_{i+1}, \mathbf{z}_i)\, p_{\psi_i}(\mathbf{z}_i | \mathbf{x}_{i+1})\, d\mathbf{z}_i,
\]
where the integrand distributions have parameters $\{\theta_i, \psi_i\}$. 
Due to the complexity and high dimensionality of the FG conformation space, 
$p_i(\mathbf{x}_i | \mathbf{x}_{i+1})$ is intractable. 
So, when designing the de-coarsening operator, instead of maximizing the log-likelihood $\log p_i(\mathbf{x}_i | \mathbf{x}_{i+1})$, we maximize the evidence lower bound (ELBO):

\[
\begin{aligned}
\log p_i(\mathbf{x}_i | \mathbf{x}_{i+1})
&\ge \;\mathbb{E}_{\,q_{\phi_i}(\mathbf{z}_i | \mathbf{x}_i,\,\mathbf{x}_{i+1})}
   \bigl[\log p_{\theta_i}(\mathbf{x}_i | \mathbf{x}_{i+1},\,\mathbf{z}_i)\bigr]
\;\\&+\;
D_{\mathrm{KL}}\Bigl(p_{\psi_i}(\mathbf{z}_i | \mathbf{x}_{i+1})
\;\big\|\;
q_{\phi_i}(\mathbf{z}_i | \mathbf{x}_i,\,\mathbf{x}_{i+1})\Bigr),
\end{aligned}
\tag{1}
\label{equ:1}
\]

where we used Jensen's inequality, and \(D_{\mathrm{KL}}\) is the Kullback--Leibler divergence, a measure of the statistical distance between two distributions.
The derived ELBO, consisting of a reconstruction term and a regularization term, can be maximized with the relevant 
\(p_{\theta_i}(\mathbf{x}_i | \mathbf{x}_{i+1}, \mathbf{z}_i)\), \(p_{\psi_i}(\mathbf{z}_i | \mathbf{x}_{i+1})\), and \(q_{\phi_i}(\mathbf{z}_i | \mathbf{x}_i, \mathbf{x}_{i+1})\)
(by using neural networks) to be illustrated in Section~\ref{sec: method}.

\medskip

Following the above inequality, we derive the \(k\)-step backmapping ELBO:

\[
\begin{aligned}
\log p(\mathbf{x}_0 | \mathbf{x}_k)
% &= 
%\log \prod_{i=0}^{k-1} p_i\bigl(\mathbf{x}_i | \mathbf{x}_{i+1}\bigr) \\[8pt] &= 
% \sum_{i=0}^{k-1} \log p_i\bigl(\mathbf{x}_i | \mathbf{x}_{i+1}\bigr)
% \\[8pt]
&\ge \sum_{i=0}^{k-1}
\Bigl(
  \mathbb{E}_{\,q_{\phi_i}(\mathbf{z}_i | \mathbf{x}_i,\,\mathbf{x}_{i+1})}
    \bigl[\log p_{\theta_i}\bigl(\mathbf{x}_i | \mathbf{x}_{i+1},\,\mathbf{z}_i\bigr)\bigr]
\\[8pt]
&
  +\,D_{\mathrm{KL}}
    \bigl(
      p_{\psi_i}\bigl(\mathbf{z}_i | \mathbf{x}_{i+1}\bigr)
      \,\big\|\,
      q_{\phi_i}\bigl(\mathbf{z}_i | \mathbf{x}_i,\,\mathbf{x}_{i+1}\bigr)
    \bigr)
\Bigr).
\end{aligned}
\tag{2}
\label{eq:2}
\]
where the new ELBO is split into \(k\) independent components, each consisting of a reconstruction term 
and a regularization term. Due to the terms' independence, we can design \(k\) models, each trying to reconstruct 
\(\mathbf{x}_i\) from \(\mathbf{x}_{i+1}\). We can thus set up \(k\) independent optimization processes to approximate the ELBO of 
\(\log p(\mathbf{x}_0 | \mathbf{x}_k)\) for a \(k\)-step scheme with the following apparent advantages:

\begin{itemize}
    \item \textbf{Optimized utilization of processing power and RAM:}
    Even with high-performance computing and modern GPUs, the demands for RAM are high
    in the case of a single model restoring the FG resolution from a CG structure with very few beads, particularly due to expensive operations on graphs.
    % For example, using a state-of-the-art general backmapping model~\cite{6}, reconstructing a 181-residue protein from 10 beads requires a fairly complex model with expensive message passing and pooling operations, limiting us to shallower models. Introducing an intermediate CG scale with $\sim 181$ beads can make each of the two sub-tasks less memory expensive and computationally intensive.

    \item \textbf{Varied reconstruction strategies:}
    We get to look at the $k$ tasks separately and introduce variations to the network architectures,
    the inductive bias, or even the objective functions. This allows adjusting the model to properties aligned with each scale independently and, likely, more efficiently.

    \item \textbf{Enhanced detection of errors:}
    A lumped single-step model lacks the ability to reveal model or implementation errors, unlike the $k$-step model that progressively advances to show all details including errors.

    \item \textbf{Reduced and smoother search spaces:}
    It is less likely for each task to be trapped in local minima during training
    because the search space for each problem is reduced, and it can be explored more thoroughly and differently.
\end{itemize}

Overall, $k$-step backmapping, essentially a ``divide and conquer'' strategy, 
divides a challenging task into $k$ relatively easier sub-tasks and then ``conquers'' them.
However, the cost of the extra human and computing resources in RAM and floating-point operations
in executing the $k$ optimization processes must be well balanced with the gain of accuracy
of the $k$-step backmapping scheme.
Choosing a proper $k$ for the ``most gain and least pain'' requires a given task and, in our work,
schemes with $k=2$, i.e., 2-step backmapping schemes,
allow demonstration of the schemes' features.

\section{Method}
\label{sec: method}

The theoretical framework, especially the ELBO derivation, has indicated the choice of
architecture and form of the objective function. After elaborating on the design of 1-step schemes,
including successful implementations by other groups, we describe our multistep method.

\subsection{Relevant architectures}
\label{sec:relevant_arch}

The variational inference and the derivation of the ELBO evoke the use of a conditional
Variational Autoencoder (c-VAE) \cite{31}, an extension of the original VAE \cite{32}.
The c-VAEs are used in semi-supervised learning for approximating distributions
conditioned on some auxiliary variables. In this case, the low-resolution CG representation
we aim to reconstruct is considered the auxiliary variable.
Following the examples of recent backmapping solutions with c-VAEs \cite{6,11} we employ neural networks
\(p_{\theta_i}(\mathbf{x}_i | \mathbf{x}_{i+1}, \mathbf{z}_i)\), \(p_{\psi_i}(\mathbf{z}_i | \mathbf{x}_{i+1})\), \(q_{\phi_i}(\mathbf{z}_i | \mathbf{x}_i, \mathbf{x}_{i+1})\)
with parameters \(\{\theta_i, \psi_i, \phi_i\}\) to approximate the corresponding conditional distributions.
They are the decoder, prior, and encoder, respectively, following the conventional terminology.
The encoder approximates a lower dimensional posterior distribution \((\mathbf{x}_i, \mathbf{x}_{i+1}) \to \mathbf{z}_i\),
shaping it as a multivariate Gaussian with learnable mean vector and covariance matrix.
Similarly, during the inference stage---when the higher resolution information is unknown---the prior
approximates the posterior distribution \(\mathbf{x}_{i+1} \to \mathbf{z}_i\).
With the KL divergence in Eq.~\ref{equ:1}, we make sure that the encoder and the prior
give rise to similar latent representations. Finally, a sample of the latent space
passes through the decoder to complete the reconstruction \((\mathbf{x}_{i+1}, \mathbf{z}_i) \to \mathbf{x}_i\).

Additionally, molecular representations necessitate graph-based algorithms when designing the
encoder, prior, and decoder. 
% At a given resolution level \(i\), the conformation can be described
% by a graph \(G_i = (V_i, E_i)\). At the FG level \((i=0)\) the node features can be the atomic types,
% at the RBCG level they could be the amino acid types, while the edge features can be determined
% by the inter-bead distances. Graphs encapsulate more information than raw coordinates and ensure
% equivariance of the operator under translations and rotations, a necessary property given
% the E(3) equivariance of the coarsening operator 
% \(\Gamma_{k-1} \circ \cdots \circ \Gamma_0\) \cite{6}.
% A well-designed encoder, decoder, and prior must be constructed from the appropriate graph architectures.
In order to accurately predict interatomic forces and other physical and geometric properties
at the quantum, atomic, and molecular levels, researchers \cite{33,34,35,36,37,38} have developed several
neural models for graph-structured data, and their architectures proved versatile across
diverse tasks \cite{39}. For instance, Message Passing Neural Networks (MPNNs) \cite{33} predicted
the local minima in the energy landscape of molecules \cite{40} as well as the primary structures
of conformations of alpha carbon traces \cite{41}, while the continuous-filter convolution network SchNet
enabled the correct derivation of CG force fields \cite{42}. These same architectures can function
as the backbone for the various components of the c-VAE.

\subsection{Backmapping solutions and 1-step baseline}
\label{sec:one_step_baseline}

Our reference 1-step scheme is \textsc{CGVAE} \cite{6}, which uses a c-VAE and the variational inference
framework with \(k=1\), and such a choice has a compelling rationale.
First, this model, equivariant under translations and rotations, is probabilistic, and thus the same
input CG can generate diverse---while accurate and natural---FG outputs.
Unlike some of its predecessors \cite{8,9}, it is not limited to condensed phase systems.
Second, among the models handling CG representations of varying resolutions, this model is much coarser
than the typical MARTINI or RBCG, and is among the most accurate.
The encoder and prior consist of continuous convolution steps with SchNet-like models \cite{37}.
In particular, they use MPNNs \cite{33} with radial basis transformations \cite{34} to process information
at the FG and CG level, as well as pooling operations to map numerical representations
from the FG to the CG space.
Similarly, the decoder is designed to process the CG conformation and samples of the latent space
through convolutions and lifting operations that ensure equivariance.
The model's objective function is based on the ELBO of Eq.~\ref{equ:1}, in which the reconstruction term
\(\mathbb{E}_{q_{\phi_0}(\mathbf{z}_0 | \mathbf{x}_0, \mathbf{x}_1)} \log p_{\theta_0}(\mathbf{x}_0 | \mathbf{x}_1, \mathbf{z}_0)\)
of the right-hand side is computed as a weighted sum of the MSD between the true \(\mathbf{x}_0\)
and the generated \(\hat{\mathbf{x}}_0\), and a term penalizing incorrect bond lengths
of adjacent atoms in the generated graph.

\medskip
Another model of interest is \textsc{GenZProt} \cite{11}, which, resembling \textsc{CGVAE} with subtle differences,
is uniquely suited for specific backmapping tasks.
Like \textsc{CGVAE}, \textsc{GenZProt} employs a c-VAE framework and leverages GNNs to encode structural information,
incorporating node embeddings with both atomic and residue-type information.
However, it differs by adopting an internal coordinate-based representation, predicting bond lengths,
bond angles, and torsion angles instead of Cartesian coordinates.
This approach preserves physical plausibility and chemical connectivity, a crucial feature for proteins.
Additionally, \textsc{GenZProt} introduces a hierarchical message-passing scheme that operates on three levels:
atom--atom, atom--residue, and residue--residue interactions, capturing both short and long-ranged dependencies.
Its key strength lies in its transferability, achieved through training on diverse protein structures,
enabling it to generalize across a wide range of protein structures and environments.
While \textsc{GenZProt} excels in reconstructing FG structures from RBCG mappings, it is not compatible with very coarse
mappings, limiting the scope of its applications.
Nevertheless, its ability to provide chemically accurate and reliable FG structures makes it
an excellent tool we see in parallel to our proposed multistep schemes.

\subsection{\texorpdfstring{$k$}{k}-step schemes}
\label{sec:k_step_schemes}

We describe how to build the 2-step scheme, which can be scaled to any multistep scheme.
Starting from a very coarse representation $\mathbf{x}_2$, we use a c-VAE model to predict
the coordinates of the alpha carbon (C\(\alpha\)) traces of the molecule $\mathbf{x}_1$,
instead of the atomistic representation.
In the absence of a real chemical structure with covalent bonds between atoms,
but artificial bond--graph edges between beads or C\(\alpha\) atoms,
we use a \textsc{CGVAE}, because it is designed to work with arbitrary CG mapping protocols.
During the second step, we restore the atomistic resolution $\mathbf{x}_0$
given the generated RBCG representation $\mathbf{x}_1$.
To take advantage of the uniformity of amino acids as protein building blocks,
as well as the chemical transferability of the RBCG $\rightarrow$ FG reconstruction \cite{11},
we choose \textsc{GenZProt} for this task.
Because proteins can be in multiple protonation states, we ignore hydrogens
and focus on reconstructing heavier atoms.
In total, we combine \textsc{CGVAE}'s flexibility with \textsc{GenZProt}'s chemical specificity.
Both models could be replaced by similar c-VAE models if they are more accurate
or suitable for each of the two tasks.

\begin{algorithm}[tb]
\caption{Backmapping a protein in $k$ steps (inference phase)}
\label{algo:k_step_backmapping}
\begin{algorithmic}
   \STATE {\bfseries Input:} $\Gamma_0, \Gamma_1, \dots, \Gamma_{k-1}$; 
          trained networks 
          $p_{\theta_i}(\mathbf{x}_i | \mathbf{x}_{i+1}, \mathbf{z}_i)$ 
          and 
          $p_{\psi_i}(\mathbf{z}_i | \mathbf{x}_{i+1})$ 
   \STATE $\mathbf{x}_k \gets (\Gamma_{k-1} \circ \cdots \circ \Gamma_0)(\mathbf{x}_0)$
   \FOR{$i=k-1$ \textbf{down to} $0$}
       \STATE Sample $\mathbf{z}_i \sim p_{\psi_i}\bigl(\mathbf{z}_i | \mathbf{x}_{i+1}\bigr)$
       \STATE $\widehat{\mathbf{x}}_i \sim 
              p_{\theta_i}\bigl(\mathbf{x}_i | \mathbf{x}_{i+1}, \mathbf{z}_i\bigr)$
       \STATE $\mathbf{x}_i \gets \widehat{\mathbf{x}}_i$
   \ENDFOR
\end{algorithmic}
\end{algorithm}

Given $\mathbf{x}_0, \mathbf{x}_1, \mathbf{x}_2$, we separately and simultaneously design two operators,
one for each step, namely the $\mathbf{x}_2 \rightarrow \mathbf{x}_1$ and $\mathbf{x}_1 \rightarrow \mathbf{x}_0$ reconstructions.
Thus, the search space of each optimization task is simplified and can be explored diversely.
For example, during RBCG $\rightarrow$ FG, we penalize unphysical bond, angle, or dihedral magnitudes,
incorporating physics and chemistry insights into the training,
while using amino acid types as node embeddings.
Moreover, we relax the computational and memory limitations; each model can be deeper,
allowing it to focus on more details for better optimization,
as escapes from local minima are unrealistic structures.
Their synergistic impact yields a more precise operator.
% A 3-step scheme is designed in a similar way, with two successive \textsc{CGVAE} models mapping a UCG conformation into a chain of \(\mathrm{C_\alpha}\) atoms, followed by \textsc{GenZProt} to finally restore atomistic resolution. 
Figure~\ref{fig:f_3} illustrates the more general $k$-step reconstruction process, while Algorithm~\ref{algo:k_step_backmapping} outlines the steps.

\subsection{Evaluation Metrics}
\label{sec:evaluation-metrics}

We adapt the following metric to assess the performance of a given scheme:

\begin{itemize}
    \item \textbf{Reconstruction Accuracy:} The accuracy is quantified with the RMSD between the true $\mathbf{x}_0$ and the generated $\widehat{\mathbf{x}}_0$ FG structures computed using only the heavy (non-hydrogen) atom coordinates:
    \[
    \mathrm{RMSD}(\mathbf{x}_0,\widehat{\mathbf{x}}_0) = \|\mathbf{x}_0 - \widehat{\mathbf{x}}_0\|_{2},
    \]
    averaged over multiple samples. We ignore hydrogens to account for the multiple protonation states of proteins.
    
    \item \textbf{Sample Quality:} We quantify how accurately the bond-graph is restored during backmapping. We compute the Graph Edit Distance \cite{43} of the true $G$ and generated $\widehat{G}$ bond graph, and divide it by the number of edges, to normalize the value:
    \[
    \lambda(G, \widehat{G}) = \frac{\mathrm{GED}(G,\widehat{G})}{|E|},
    \]
    given the bond graphs $G = (V, E)$ and $\widehat{G} = (\widehat{V}, \widehat{E})$. A lower GED means that it takes fewer edit operations, i.e., vertex/edge insertions/deletions/substitutions, to transform one graph into another, accounting for the total number of edges $|E|$ for normalization purposes. 
    % A lower value of sample quality corresponds to a more accurate reconstruction of the bond graph.
    
    \item \textbf{Steric Clash Score:} A steric clash occurs when two non-bonded atoms of the protein are unnaturally close to each other --- such a positioning would normally lead to a strong repulsion force. When it comes to backmapping, the fewer the steric clashes, the more realistic the resulting conformation. If the distance between any two heavy (non-hydrogen) sidechain atoms from neighboring residues falls within a threshold distance, then we report a steric clash. The threshold distance is set to 1.2\,\AA, as in \cite{11}. We compute the percentage of residues with a steric clash. 
    % We aim for reconstructions with as low of a steric clash score as possible.
\end{itemize}

For each setup, we carry out statistical analysis, and report the mean and standard deviation of each metric by performing independent experiments. We also conduct visual assessment of the secondary structure of reconstructed conformations by examining the Ramachandran plot \cite{44} to illustrate the distributions of the main backbone dihedrals. 
% More specifically, for each residue we compute the dihedral angles measuring the rotation around the N--C$\alpha$ and C$\alpha$--C bonds, denoted by $\phi$ and $\psi$, respectively. We plot the distribution of different $(\phi, \psi)$ combinations of the FG dataset and the reconstructed FG structures for different molecules, CG bead sizes, and backmapping schemes. 
A strong match between the Ramachandran plots of the true and generated FG structures signifies a more accurate backmapping scheme.

\section{Implementation}
\label{sec:implementation}

We apply the 2-step iterative scheme to two proteins and compare it with \textsc{CGVAE}, our 1-step baseline. We choose \textsc{CGVAE} because it is probabilistic, equivariant under translations and rotations, and the most accurate model applied to UCG protein representations.

\subsection{Data}
\label{subsec:data}

The first molecule we apply the models to is the eukaryotic translation initiation factor 4E, namely the protein eIF4E (PDB: 4TPW \cite{45}). The structure we use has 181 residues, and long trajectories of the molecule can be found in \cite{46}. Also, we choose mappings with very few beads, specifically \(n_2 = 10, 19, 50\). Considering the number of residues in the molecule, this corresponds to average CG bead sizes \(\rho = 18.10, 9.53, 3.62\), respectively.  
The second molecule is a structural ensemble of a Nuclear Localization Signal (NLS 99–140) peptide \cite{47}, labeled as PED000151 in the Protein Ensemble Database \cite{48}. It has 48 residues and was reduced to \(n_2 = 5, 8\) beads, leading to average CG bead sizes of \(\rho = 9.60, 6.00\), respectively. In contrast, previous works \cite{6,11,18} have reported CG bead sizes ranging from 0.33 to 1.67.%, as indicated in Fig.~\ref{fig:cg-sizes}.

We choose two proteins with different structural and dynamical features. eIF4E is globular, compact, with a more limited range of conformations. On the other hand, PED00151 is an Intrinsically Disordered Protein (IDP), meaning its three-dimensional structure is not stable under normal conditions. 
%In Fig.~\ref{fig:radius}, we plot the distributions of the radius of gyration \(R_g\) \cite{49} of the two proteins over the course of \(1\,\mu\text{s}\) equilibrium simulations, showcasing their flexibility or lack thereof.
The radius of gyration \(R_g\) \cite{49} of the two proteins over the course of \(1\,\mu\text{s}\) equilibrium simulations shows their flexibility or lack thereof. More specifically, with a mean \(R_g\) of 1.627 nm and a standard deviation of 0.027, as opposed to a mean of 2.152 nm and a larger standard deviation of 0.285, eIF4E is clearly more compact than the flexible PED00151.

% \begin{figure}
%     \centering
%     \includegraphics[width=0.65\linewidth]{Figures/Picture5.png}
%     \caption{Radius of gyration distribution of the two proteins we test the schemes on. The variance differences indicate eIF4E is more compact, while PED00151 is more flexible.}
%     \label{fig:radius}
% \end{figure}

\subsection{Training}
\label{subsec:training}

Prior to model training, we collect three independent \(1\,\mu\text{s}\) simulations of eIF4E from \cite{46}, leading to a total of 3000 FG conformations. Similarly, we collect 9\,746 simulation frames of the PED000151e000 ensemble from \cite{11}. We use \textsc{AutoGrain} \cite{6} to form various CG operators:
\[
\mathbf{x}_0 \!\to\! \mathbf{x}_2,\;
\mathbf{x}_0 \!\to\! \mathbf{x}_3,\;
\mathbf{x}_1 \!\to\! \mathbf{x}_2,\;
\mathbf{x}_1 \!\to\! \mathbf{x}_3,\;
\text{and}\;
\mathbf{x}_2 \!\to\! \mathbf{x}_3 .
\]
\textsc{AutoGrain}'s objective function prevents distant beads in a high-resolution representation from being assigned to the same low-resolution CG bead. This way, we create geometry-conserving CG mapping operators with the specified number of beads. Moreover, we form the corresponding \(\mathbf{x}_0 \!\to\! \mathbf{x}_1\) mappings and representation \(\mathbf{x}_1\) using MDTraj \cite{50} by selecting the alpha carbon C\(\alpha\) of each residue. 
% In Fig.~\ref{fig:rep-chain} we visualize all the different resolutions.  
%Note that the 3\-step scheme is not applied to eIF4E. As we will see later in Section~4.3, the reconstruction accuracy of the 2\-step scheme falls within \(1.50\;\text{\AA}\), so we do not see the need to apply a 3\-step scheme for further refinement. 
By visual inspection, we claim that knowledge of the coarser representation does not provide more information than the RBCG representation. This way we can invoke the Markov property and take advantage of conditional independence, as described in Section~\ref{sec: label}. For a more detailed look at the CG representations, plots are provided in Appendix~\ref{sec:cg_representations}.

We train multiple \textsc{CGVAE}-like models with a train–validation–test split (80-10-10). In a similar way, \textsc{GenZProt}, trained on a variety of proteins \cite{11}, is fine-tuned on eIF4E and PED000151 structures with an additional split on the corresponding datasets. We set a number of epochs with early stopping and an adaptive learning-rate scheduler, in case the validation-loss improvement stagnates. To search the hyperparameter space efficiently and extensively we use the Weights and Biases software \cite{51} with the hyperband algorithm \cite{52}. 
For the 2-step scheme, the batch size was set to 2. Due to memory overload, the batch size for the 1-step scheme was set to 1 with gradient accumulation \cite{53}. 
You can find detailed tables with the hyperparameters corresponding to each condition in Appendix~\ref{sec:hyperparameters}.

We train our models on a high-performance computing cluster. For 1-step backmapping, we use two NVIDIA A100 Tensor Core 80\,GB GPUs with an Intel Haswell CPU due to the heavy memory and computational load, mostly originating from operations on graphs. We incorporate data parallelization \cite{54} in the original PyTorch \cite{55} code. For 2-step backmapping, the memory and computational load from the UCG $\rightarrow$ RBCG reconstruction is much lower, so we only use one NVIDIA Tesla K80 24\,GB GPU with an Intel Ice Lake CPU for each step.

\subsection{Results}
\label{subsec:results}

\begin{figure}[h] 
  \centering
  \includegraphics[width=\linewidth]{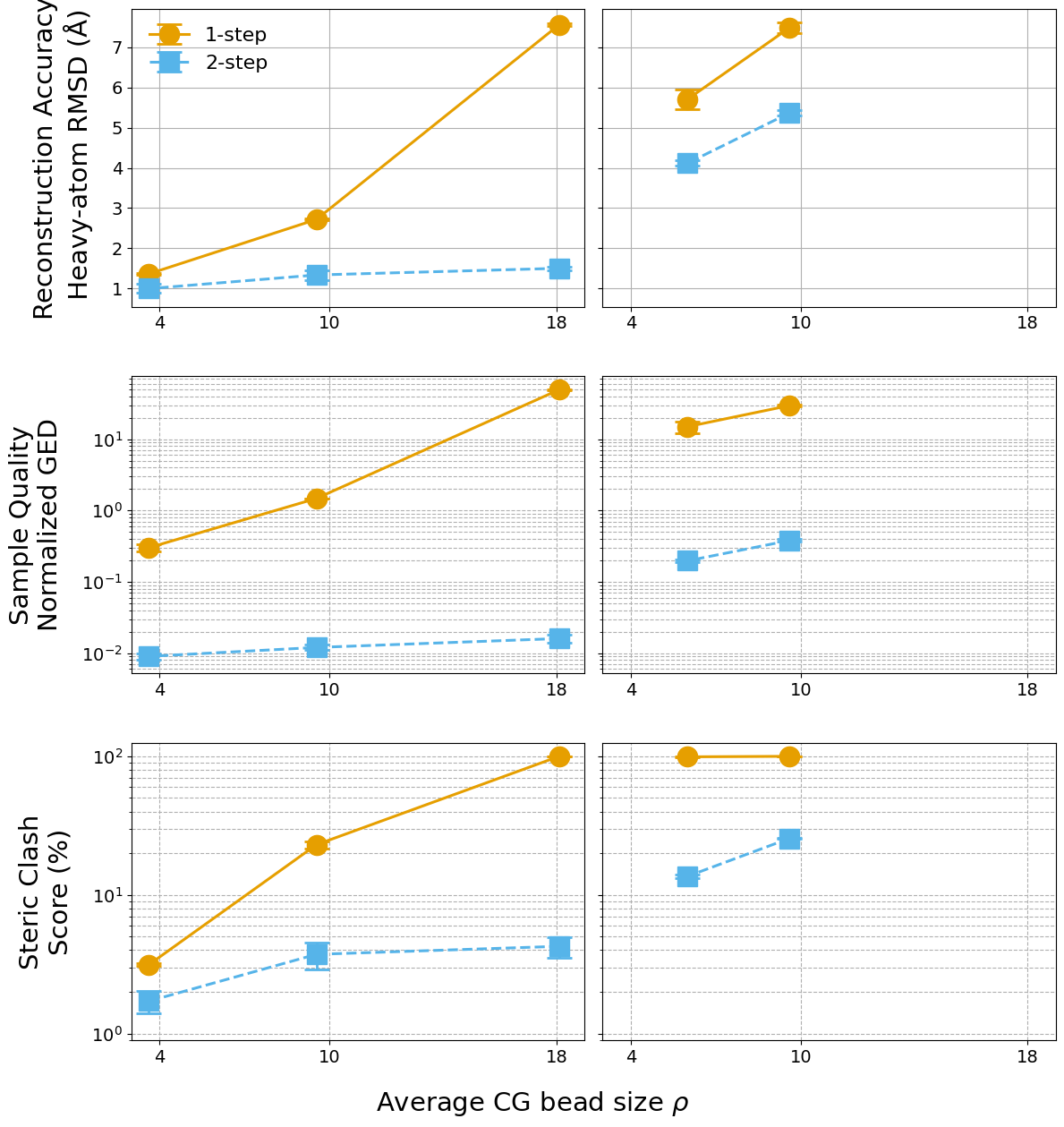} % ← fig7.pdf / fig7.png in the same folder
  \caption{%
    The results for eIF4E (left) and PED00151 (right). The vertical axis in the bottom two rows is shown on a logarithmic scale.%
  }
  \label{fig:eif4e-summary}
\end{figure}

To evaluate the performance of our proposed iterative backmapping schemes, we conduct a comparative analysis against the 1-step \textsc{CGVAE} baseline using two proteins of different structural characteristics: the globular protein eIF4E and the intrinsically disordered protein PED00151. We apply the 2-step backmapping scheme to both proteins. 
%and additionally evaluate a 3-step scheme for PED00151 owing to its higher conformational heterogeneity.  
The results are reported in Tables~\ref{tab:eif4e_results} and~\ref{tab:ped00151_results} and are illustrated in Fig.~\ref{fig:eif4e-summary} --- note that the vertical axis for sample quality and steric-clash scores is in logarithmic scale. We record the mean and standard deviation from three experiments with different random seeds. We observe consistent improvements across multiple reconstruction metrics when employing multistep schemes.

% ----------------------- eIF4E table (smaller font) -----------------------
% ------------ eIF4E --------------------------------------------------------
\begin{table*}[t]
\caption{Results on the protein eIF4E.}
\label{tab:eif4e_results}
\vskip 0.15in
\begin{center}
\begin{small}
\begin{sc}
\begin{tabular}{llccc}
\toprule
Metric & Scheme & 18.10 (10) & 9.53 (19) & 3.62 (50) \\
\midrule
\multirow{2}{*}{Reconstruction Accuracy RMSD (\AA)} 
    & 1-step & 7.563 $\pm$ 0.027 & 2.718 $\pm$ 0.018 & 1.364 $\pm$ 0.024 \\
    & 2-step & 1.503 $\pm$ 0.045 & 1.337 $\pm$ 0.123 & 0.997 $\pm$ 0.111 \\
\midrule
\multirow{2}{*}{Sample Quality (normalized GED)} 
    & 1-step & 49.562 $\pm$ 0.437 & 1.476 $\pm$ 0.009 & 0.301 $\pm$ 0.032 \\
    & 2-step & 0.016 $\pm$ 0.002 & 0.012 $\pm$ 0.001 & 0.009 $\pm$ 0.001 \\
\midrule
\multirow{2}{*}{Steric Clash Score (\%)} 
    & 1-step & 99.999 $\pm$ 0.001 & 23.030 $\pm$ 1.287 & 3.144 $\pm$ 0.057 \\
    & 2-step & 4.264 $\pm$ 0.728 & 3.733 $\pm$ 0.809 & 1.716 $\pm$ 0.315 \\
\bottomrule
\end{tabular}
\end{sc}
\end{small}
\end{center}
\vskip -0.1in
\end{table*}

% ------------ PED00151 -----------------------------------------------------
\begin{table*}[t]
\caption{Results on the protein PED00151.}
\label{tab:ped00151_results}
\vskip 0.15in
\begin{center}
\begin{small}
\begin{sc}
\begin{tabular}{llcc}
\toprule
Metric & Scheme & 9.60 (5) & 6.00 (8) \\
\midrule
\multirow{2}{*}{Reconstruction Accuracy RMSD (\AA)} 
    & 1-step & 7.488 $\pm$ 0.129 & 5.707 $\pm$ 0.244 \\
    & 2-step & 5.368 $\pm$ 0.059 & 4.118 $\pm$ 0.066 \\
\midrule
\multirow{2}{*}{Sample Quality (normalized GED)} 
    & 1-step & 29.665 $\pm$ 0.753 & 15.000 $\pm$ 2.799 \\
    & 2-step & 0.330 $\pm$ 0.015 & 0.197 $\pm$ 0.010 \\
\midrule
\multirow{2}{*}{Steric Clash Score (\%)} 
    & 1-step & 99.970 $\pm$ 0.032 & 99.099 $\pm$ 0.799 \\
    & 2-step & 25.611 $\pm$ 0.258 & 13.631 $\pm$ 0.474 \\
\bottomrule
\end{tabular}
\end{sc}
\end{small}
\end{center}
\vskip -0.1in
\end{table*}

\paragraph{Quantitative Evaluation}%
We assess reconstruction accuracy using heavy-atom RMSD, sample quality via normalized GED, and structural plausibility through the steric clash score.  
For eIF4E, the 2-step scheme significantly reduces the RMSD, especially at the highest CG bead size (10 beads, \(\rho = 18.10\)), where the RMSD drops from 7.56 Å (1-step) to 1.50 Å (2-step). Improvements are also consistent at finer resolutions (\(\rho = 9.53\) and \(3.62\)), with RMSD values falling below 1.00 Å for the finest CG.  
This performance gain confirms that intermediate CG representations help guide the reconstruction towards more accurate FG configurations. The sample quality improves by more than an order of magnitude across all resolutions; for instance, at \(\rho = 18.10\) it decreases from 49.56 to 0.016. The steric-clash score demonstrates a pattern similar to the other metrics, indicating that intermediate resolutions lead to more chemically plausible reconstructions.

For PED00151, the more flexible IDP, results show a clear but more moderate improvement with the 2-step scheme. The RMSD decreases from 7.49 Å (1-step) to 5.37 Å (2-step) at \(\rho = 9.60\) and from 5.71 Å to 4.12 Å at \(\rho = 6.00\). These gains are less dramatic than in eIF4E, highlighting the inherent challenges in backmapping disordered proteins. 
%Still, 3-step back-mapping offers further improvements \dots, supporting our hypothesis that finer intermediate resolutions can alleviate the high variability of disordered structures.  
In terms of steric clashes, PED00151 shows extremely high scores with the 1-step baseline (nearly 100 \%), indicating severe structural artifacts. The 2-step scheme dramatically reduces these values to 25.61 \% and 13.63 \%, respectively, showing that iterative refinement significantly enhances physical realism.

\begin{figure}[h] 
  \centering
  \includegraphics[width=\linewidth]{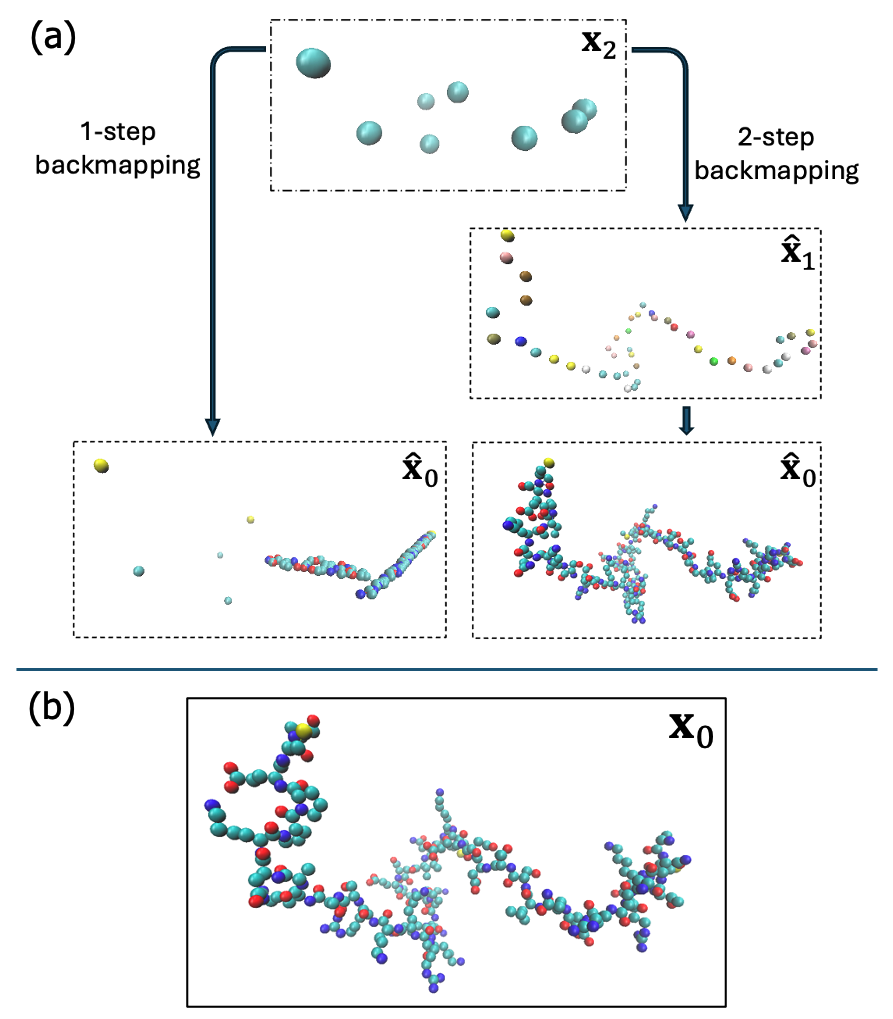}
  \caption{%
    (a) Starting from $\mathbf{x}_2$ with $n_2=8$, we restore the FG representation $\mathbf{x}_0$ of PED00151 using 1-step and 2-step schemes. (b) The ground truth $\mathbf{x}_0$.
  }
  \label{fig:PED_examples}
\end{figure}

Notice that, unlike sample quality and steric-clash score, reconstruction accuracy had a more modest improvement. This is because normalized GED and steric-clash score percentages are structurally binary, whereas RMSD is continuous. Even if atom positions are slightly off in a way that still contributes to RMSD, correcting just enough to form correct or remove incorrect bonds dramatically reduces GED. That is why GED can improve significantly even when RMSD does not shift much. An example is illustrated in Fig.~\ref{fig:PED_examples}, where the modest improvement of RMSD$(\mathbf{x}_0, \widehat{\mathbf{x}}_0)$ from 5.64 Å for 1-step to 3.65 Å for 2-step actually leads to a much more realistic structure. This example also raises questions about the reliability of RMSD as a metric often used in the literature to quantify structural deviation.

\paragraph{Secondary Structure Recovery}%
We further validate back-mapping quality by analyzing the Ramachandran plots of reconstructed FG structures, shown in Appendix~\ref{sec:secondary} for more clarity. For eIF4E, the 2-step scheme yields dihedral-angle distributions that closely match the native FG dataset, particularly at \(\rho = 3.62\). In contrast, the 1-step scheme fails to reproduce realistic backbone angles at high \(\rho\), indicating distorted or unnatural secondary structures. This trend highlights the necessity of using intermediate resolutions to preserve features such as alpha helices and beta sheets during reconstruction.

For PED00151, even the 2-step scheme produces broader and noisier dihedral angle distributions than those of the reference FG dataset. This aligns with the nature of IDPs, where the high entropy of conformational states imposes a hard limit on reconstruction precision.

\paragraph{Importance of Bead Size and Molecule Properties}%
Our experiments also reveal that the benefits of multistep schemes are more pronounced at higher average CG bead sizes. At lower \(\rho\) and for more compact molecules, 1-step schemes are capable of achieving decent reconstructions because the task is less ambiguous. For instance, for eIF4E at \(\rho = 3.62\) all three metrics approach their ideal values, and the additional step in the 2-step scheme yields only modest improvements.  
At coarser resolutions, however, single-step models struggle owing to their limited capacity and the computational burden of restoring high-resolution detail from minimal input. Hardware constraints further emphasize this issue: because of GPU-memory limits we could train only 1-step models with shallow architectures (e.g., batch size = 1, encoder-convolution depth = 1), resulting in a restricted search space. By contrast, our 2-step scheme enables deeper networks for each stage, leading to more thorough optimization and a lower risk of convergence to sub-optimal minima.

\section{Conclusion, Limitations, and Future Work}
We present a practical framework for improving backmapping from ultra-coarse-grained (UCG) protein representations using multistep generative models. By decomposing the reconstruction into smaller, intermediate steps, our method enables deeper, more specialized networks at each resolution level—reducing memory bottlenecks and improving accuracy, especially for high CG bead sizes. Compared to single-step methods, multistep schemes offer not only better reconstruction metrics but also practical benefits such as modular design, improved convergence, and easier debugging. 

While 2-step backmapping improves structural realism over direct baselines, challenges remain, especially for IDPs, where coarse-grained representations carry less recoverable information. In future work, we aim to implement 3-step schemes, investigate more thoroughly the effect of inherent flexibility on backmapping success, as well as explore different strategies to accurately reconstruct disordered proteins or embed additional information into the reconstruction process when targeting flexible residues.

\section*{Software and Data}
The eIF4E data used for training, validation and testing can be requested from the authors of \cite{46}. The PED000151 data can be found in the Protein Ensemble Database \cite{48}. 

The network architectures are adapted from \cite{6,11} and the relevant code can be found in the following publicly available repositories: \url{https://github.com/wwang2/CoarseGrainingVAE} (\textsc{CGVAE}) and \url{https://github.com/learningmatter-mit/GenZProt} (\textsc{GenZProt}). To train these models, you can access detailed tables with the hyperparameters in Appendix~\ref{sec:hyperparameters}.

% The network architectures are adapted from \cite{6,11} and the relevant code can be found in their publicly available repositories: \url{https://github.com/wwang2/CoarseGrainingVAE} (\textsc{CGVAE}) and \url{https://github.com/learningmatter-mit/GenZProt} (\textsc{GenZProt}). To train these models, you can access detailed tables with the hyperparameters in Appendix~\ref{sec:hyperparameters}.

\section*{Acknowledgements}
The authors would like to thank the Institute of Advanced Computational Science (IACS) at Stony Brook University for providing the high-performance computing cluster Seawulf where the model training and testing took place.

\bibliographystyle{icml2025}

\newpage
\appendix
\onecolumn

\section{Table of Notations}
\begin{table}[H]
\caption{Explanation of symbols and notation in the paper.}
\label{tab:symbols}
\vskip 0.15in
\begin{center}
\begin{small}
%\begin{sc}
\begin{tabular}{ll}
\toprule
\textbf{Symbol} & \textbf{Meaning} \\
\midrule
$\mathbf{x}_i$ & Coordinates at step $i$ \\
$n_i$          & Number of particles at step $i$ \\
$\mathbf{z}_i$ & Latent space variable at step $i$ \\
$\Gamma_i$     & Coarsening operator at step $i$ \\
$p_{\theta_i}(\mathbf{x})$ & Distribution of $\mathbf{x}$ with parameters $\theta_i$ \\
$\widehat{\mathbf{x}}_i$ & Predicted coordinates at step $i$ \\
$p(\mathbf{x}|\mathbf{y})$ & Conditional probability of $\mathbf{x}$ given $\mathbf{y}$ \\
$D_{\mathrm{KL}}(p \| q)$ & Kullback–Leibler Divergence of $p$ and $q$ \\
$\rho$ & Average CG bead size in unit of residue count \\
\bottomrule
\end{tabular}
%\end{sc}
\end{small}
\end{center}
\vskip -0.1in
\end{table}

\section{Detailed Derivations}
Given two representations $\mathbf{x}_i, \mathbf{x}_{i+1}$, we derive the Evidence Lower Bound (ELBO):
\[
\begin{aligned}
\log p_i(\mathbf{x}_i | \mathbf{x}_{i+1})
&= \;\log \,\mathbb{E}_{\,q_{\phi_i}(\mathbf{z}_i | \mathbf{x}_i,\,\mathbf{x}_{i+1})}\!\Bigl[
   \tfrac{p_i(\mathbf{x}_i | \mathbf{x}_{i+1})}{\,q_{\phi_i}(\mathbf{z}_i | \mathbf{x}_i,\,\mathbf{x}_{i+1})}\Bigr]
\\[6pt]
&\ge \;\mathbb{E}_{\,q_{\phi_i}(\mathbf{z}_i | \mathbf{x}_i,\,\mathbf{x}_{i+1})}\!\Bigl[
   \log \tfrac{p_i(\mathbf{x}_i | \mathbf{x}_{i+1},\,\mathbf{z}_i)\;p_{\psi_i}(\mathbf{z}_i | \mathbf{x}_{i+1})}
            {q_{\phi_i}(\mathbf{z}_i | \mathbf{x}_i,\,\mathbf{x}_{i+1})}\Bigr]
\\[6pt]
&= \;\mathbb{E}_{\,q_{\phi_i}(\mathbf{z}_i | \mathbf{x}_i,\,\mathbf{x}_{i+1})}
   \bigl[\log p_{\theta_i}(\mathbf{x}_i | \mathbf{x}_{i+1},\,\mathbf{z}_i)\bigr]
\;+\;
\mathbb{E}_{\,q_{\phi_i}(\mathbf{z}_i | \mathbf{x}_i,\,\mathbf{x}_{i+1})}
   \!\Bigl[\log \tfrac{p_{\psi_i}(\mathbf{z}_i | \mathbf{x}_{i+1})}
                         {q_{\phi_i}(\mathbf{z}_i | \mathbf{x}_i,\,\mathbf{x}_{i+1})}\Bigr]
\\[6pt]
&= \;\mathbb{E}_{\,q_{\phi_i}(\mathbf{z}_i | \mathbf{x}_i,\,\mathbf{x}_{i+1})}
   \bigl[\log p_{\theta_i}(\mathbf{x}_i | \mathbf{x}_{i+1},\,\mathbf{z}_i)\bigr]
\;+\;
D_{\mathrm{KL}}\Bigl(p_{\psi_i}(\mathbf{z}_i | \mathbf{x}_{i+1})
\;\big\|\;
q_{\phi_i}(\mathbf{z}_i | \mathbf{x}_i,\,\mathbf{x}_{i+1})\Bigr),
\end{aligned}
\tag{1}
\label{equ:1}
\]

In the second row, we have applied Jensen's inequality.

Moreover, we can combine multiple ELBO's to compute a lower bound for $\log p(\mathbf{x}_0 | \mathbf{x}_k)$:
\[
\begin{aligned}
\log p(\mathbf{x}_0 | \mathbf{x}_k)
&= 
\log \prod_{i=0}^{k-1} p_i\bigl(\mathbf{x}_i | \mathbf{x}_{i+1}\bigr) \\[8pt] &= 
\sum_{i=0}^{k-1} \log p_i\bigl(\mathbf{x}_i | \mathbf{x}_{i+1}\bigr)
\\[8pt]
&\ge \sum_{i=0}^{k-1}
\Bigl(
  \mathbb{E}_{\,q_{\phi_i}(\mathbf{z}_i | \mathbf{x}_i,\,\mathbf{x}_{i+1})}
    \bigl[\log p_{\theta_i}\bigl(\mathbf{x}_i | \mathbf{x}_{i+1},\,\mathbf{z}_i\bigr)\bigr]
\\[8pt]
&
  +\,D_{\mathrm{KL}}
    \bigl(
      p_{\psi_i}\bigl(\mathbf{z}_i | \mathbf{x}_{i+1}\bigr)
      \,\big\|\,
      q_{\phi_i}\bigl(\mathbf{z}_i | \mathbf{x}_i,\,\mathbf{x}_{i+1}\bigr)
    \bigr)
\Bigr).
\end{aligned}
\tag{2}
\label{eq:2_detail}
\]

\newpage
\section{Secondary Structure}
\label{sec:secondary}
 The Ramachandran plots illustrating the dihedral angle distributions across the true FG and generated/backmapped structures. 
 %The contours correspond to the true distribution while the color corresponds to the probability density of the $(\phi, \psi)$ angle combinations across the backmapped structures.

\begin{figure}[H]
  \centering
  \includegraphics[width=\linewidth]{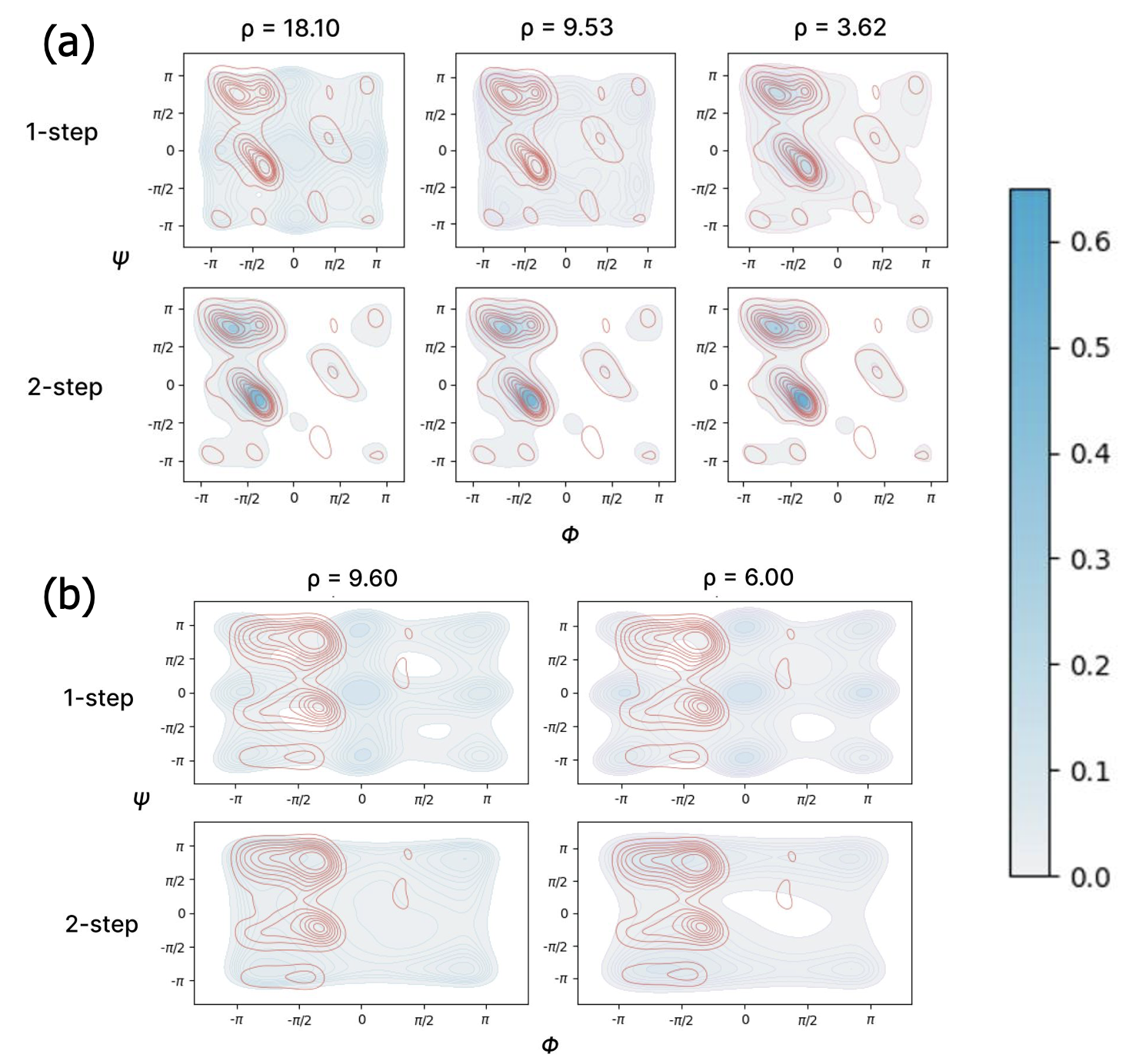} % ← fig8.pdf / fig8.png in the same folder
  \caption{%
    Ramachandran plots for different schemes and CG bead sizes: (a) eIF4E, (b) PED00151. The contours correspond to the true distribution while the color corresponds to the probability density of the $(\phi, \psi)$ angle combinations across the backmapped structures.
  }
  \label{fig:ramachandran-comparison}
\end{figure}

\section{Hyperparameter Tables}
\label{sec:hyperparameters}
We performed hyperparameter optimization using Weights \& Biases~\cite{51}. For the UCG$\rightarrow$RBCG or UCG$\rightarrow$FG reconstructions, we employed \textsc{CGVAE}~\cite{6}, tuned with the hyperband algorithm~\cite{52} to minimize validation loss. For downstream backmapping (RBCG$\rightarrow$FG), we fine-tuned a pretrained \textsc{GenZProt}, keeping its original hyperparameters~\cite{11}.

Some comments related to the implications of memory overload in 1-step models:
\begin{itemize}
    \item In 1-step schemes, we used a batch size of 1, with gradient accumulation every 2 steps, leading to an effective batch size of 2.
    \item For very coarse CG mappings, convolution depths had to be low.
    \item The node embedding dimension matches or exceeds the maximum number of mapped FG particles per bead.
\end{itemize}

\begin{table}[H]
\caption{Hyperparameters for the 1-step backmapping (UCG $\rightarrow$ FG) of eIF4E using CGVAE.}
\label{tab:hyperparams_ucg_fg_eif4e}
\vskip 0.15in
\begin{center}
\begin{small}
%\begin{sc}
\begin{tabular}{lccc}
\toprule
% \textbf{} & \multicolumn{3}{c}{\textbf{Value}} \\
% \midrule
\# of atoms (FG) & \multicolumn{3}{c}{$n_0 = 1488$} \\
Dataset size     & \multicolumn{3}{c}{3000} \\
\# of CG beads   & $n_1 = 10$ & $n_1 = 19$ & $n_1 = 50$ \\
\midrule
\textbf{Hyperparameter} & \multicolumn{3}{c}{\textbf{Value}} \\
\midrule
Edge feature dimension $K$     & 6 & 6 & 6 \\
Graph loss weight $\gamma$     & 1.303 & 29.000 & 1.188 \\
Encoder Convolution Depth      & 1 & 1 & 1 \\
Prior Convolution Depth        & 1 & 1 & 1 \\
Decoder Convolution Depth      & 1 & 4 & 4 \\
FG cutoff $d_{\mathrm{cut}}$   & 13.262 & 14.947 & 12.865 \\
CG cutoff $D_{\mathrm{cut}}$   & 31.738 & 33.064 & 23.657 \\
Node embedding dimension $F$   & 400 & 400 & 400 \\
Batch size                     & 2 & 2 & 2 \\
Learning rate                  & 0.000078 & 0.000064 & 0.000054 \\
Activation functions           & swish & swish & swish \\
Training epochs                & 100 & 100 & 100 \\
Order for multi-hop graph      & 3 & 1 & 1 \\
Regularization strength $\beta$ & 0.002 & 0.040 & 0.002 \\
Factor                         & 0.210 & 0.131 & 0.106 \\
\bottomrule
\end{tabular}
%\end{sc}
\end{small}
\end{center}
\vskip -0.1in
\end{table}

\begin{table}[H]
\caption{Hyperparameters for the first step of 2-step backmapping (UCG $\rightarrow$ RBCG) of eIF4E using CGVAE.}
\label{tab:hyperparams_ucg_rbcg_eif4e}
\vskip 0.15in
\begin{center}
\begin{small}
\begin{tabular}{lccc}
\toprule
\# of atoms (CG) & \multicolumn{3}{c}{$n_1 = 177$} \\
Dataset size     & \multicolumn{3}{c}{3000} \\
\# of CG beads   & $n_2 = 10$ & $n_2 = 19$ & $n_2 = 50$ \\
\midrule
\textbf{Hyperparameter} & \multicolumn{3}{c}{\textbf{Value}}  \\
\midrule
Edge feature dimension $K$     & 6 & 6 & 6 \\
Graph loss weight $\gamma$     & 0.8014 & 2.6083 & 4.3745 \\
Encoder Convolution Depth      & 4 & 6 & 3 \\
Prior Convolution Depth        & 4 & 6 & 3 \\
Decoder Convolution Depth      & 6 & 5 & 7 \\
FG cutoff $d_{\mathrm{cut}}$   & 40.296 & 47.771 & 43.03657 \\
CG cutoff $D_{\mathrm{cut}}$   & 81.525 & 69.453 & 115.64641 \\
Node embedding dimension $F$   & 389 & 421 & 563 \\
Batch size                     & 2 & 2 & 2 \\
Learning rate                  & 0.0001814 & 0.0001719 & 0.00010339 \\
Activation functions           & swish & swish & swish \\
Training epochs                & 30 & 30 & 30 \\
Order for multi-hop graph      & 1 & 1 & 6 \\
Regularization strength $\beta$ & 0.000199 & 0.00541 & 0.022151 \\
Factor                         & 0.53875 & 0.72021 & 0.76064 \\
\bottomrule
\end{tabular}
\end{small}
\end{center}
\vskip -0.1in
\end{table}

\begin{table}[H]
\caption{Hyperparameters for the 1-step backmapping (UCG $\rightarrow$ FG) of PED00151 using CGVAE.}
\label{tab:hyperparams_ucg_fg_ped00151}
\vskip 0.15in
\begin{center}
\begin{small}
\begin{tabular}{lcc}
\toprule
\# of atoms (FG) & \multicolumn{2}{c}{$n_0 = 376$} \\
Dataset size     & \multicolumn{2}{c}{10000} \\
\# of CG beads   & $n_1 = 5$ & $n_1 = 8$ \\
\midrule
\textbf{Hyperparameter} & \multicolumn{2}{c}{\textbf{Value}} \\
\midrule
Edge feature dimension $K$     & 6 & 6 \\
Graph loss weight $\gamma$     & 8.0601 & 1.6931 \\
Encoder Convolution Depth      & 1 & 2 \\
Prior Convolution Depth        & 1 & 2 \\
Decoder Convolution Depth      & 4 & 4 \\
FG cutoff $d_{\mathrm{cut}}$   & 14.8905 & 11.4050 \\
CG cutoff $D_{\mathrm{cut}}$   & 29.2027 & 22.0893 \\
Node embedding dimension $F$   & 450 & 400 \\
Batch size                     & 1 & 1 \\
Learning rate                  & 0.000165 & 0.0000986 \\
Activation functions           & swish & swish \\
Training epochs                & 100 & 100 \\
Order for multi-hop graph      & 1 & 1 \\
Regularization strength $\beta$ & 0.0214 & 0.0311 \\
Factor                         & 0.2294 & 0.7787 \\
\bottomrule
\end{tabular}
\end{small}
\end{center}
\vskip -0.1in
\end{table}

\begin{table}[H]
\caption{Hyperparameters for the 2-step backmapping (UCG $\rightarrow$ RBCG) of PED00151 using CGVAE.}
\label{tab:hyperparams_ucg_rbcg_ped00151}
\vskip 0.15in
\begin{center}
\begin{small}
\begin{tabular}{lcc}
\toprule
\# of atoms (FG) & \multicolumn{2}{c}{$n_1 = 46$} \\
Dataset size     & \multicolumn{2}{c}{10000} \\
\# of CG beads   & $n_2 = 5$ & $n_2 = 8$ \\
\midrule
$\textbf{Hyperparameter}$ & \multicolumn{2}{c}{$\mathbf{Value}$} \\
\midrule
Edge feature dimension $K$     & 6 & 6 \\
Graph loss weight $\gamma$     & 2.4581 & 0.7310 \\
Encoder Convolution Depth      & 4 & 5 \\
Prior Convolution Depth        & 4 & 5 \\
Decoder Convolution Depth      & 10 & 10 \\
FG cutoff $d_{\mathrm{cut}}$   & 47.9878 & 57.6470 \\
CG cutoff $D_{\mathrm{cut}}$   & 50.1431 & 100.343 \\
Node embedding dimension $F$   & 569 & 507 \\
Batch size                     & 2 & 2 \\
Learning rate                  & 0.0000975 & 0.000116 \\
Activation functions           & swish & swish \\
Training epochs                & 30 & 30 \\
Order for multi-hop graph      & 1 & 3 \\
Regularization strength $\beta$ & 0.00057 & 0.00207 \\
Factor                         & 0.639 & 0.916 \\
\bottomrule
\end{tabular}
\end{small}
\end{center}
\vskip -0.1in
\end{table}

\newpage
\section{Comparison of CG Bead Sizes}
As indicated in previous sections, this work explores reconstruction of UCG representations, unlike 
\begin{figure}[H]
    \centering
    \includegraphics[width=0.8\linewidth]{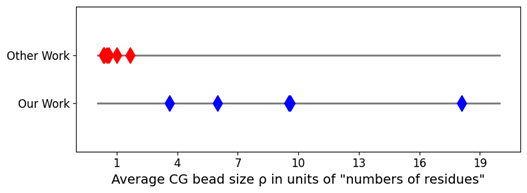}
    \caption{The CG bead sizes \(\rho\) in previous papers are much smaller— and closer to \(\rho = 1.00\)—than the ones we try in our work.}
    \label{fig:cg-sizes}
\end{figure}

\section{Radius of Gyration}
\begin{figure}[H]
    \centering
    \includegraphics[width=0.5\linewidth]{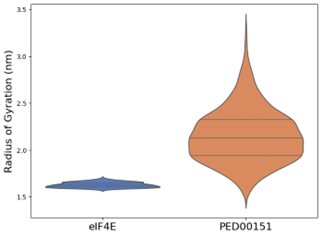}
    \caption{Radius of gyration distribution of the two proteins we test the schemes on. The difference in variance indicate eIF4E is more compact, while PED00151 is more flexible.}
    \label{fig:radius}
\end{figure}

\newpage
\section{CG Representations}
\label{sec:cg_representations}

Using \textsc{AutoGrain} \cite{6} and MDTraj \cite{50} we develop the various CG mapping operators and apply them to the FG datasets used in the experiments. 
\begin{figure}[H]
    \centering
    \includegraphics[width=0.60\linewidth]{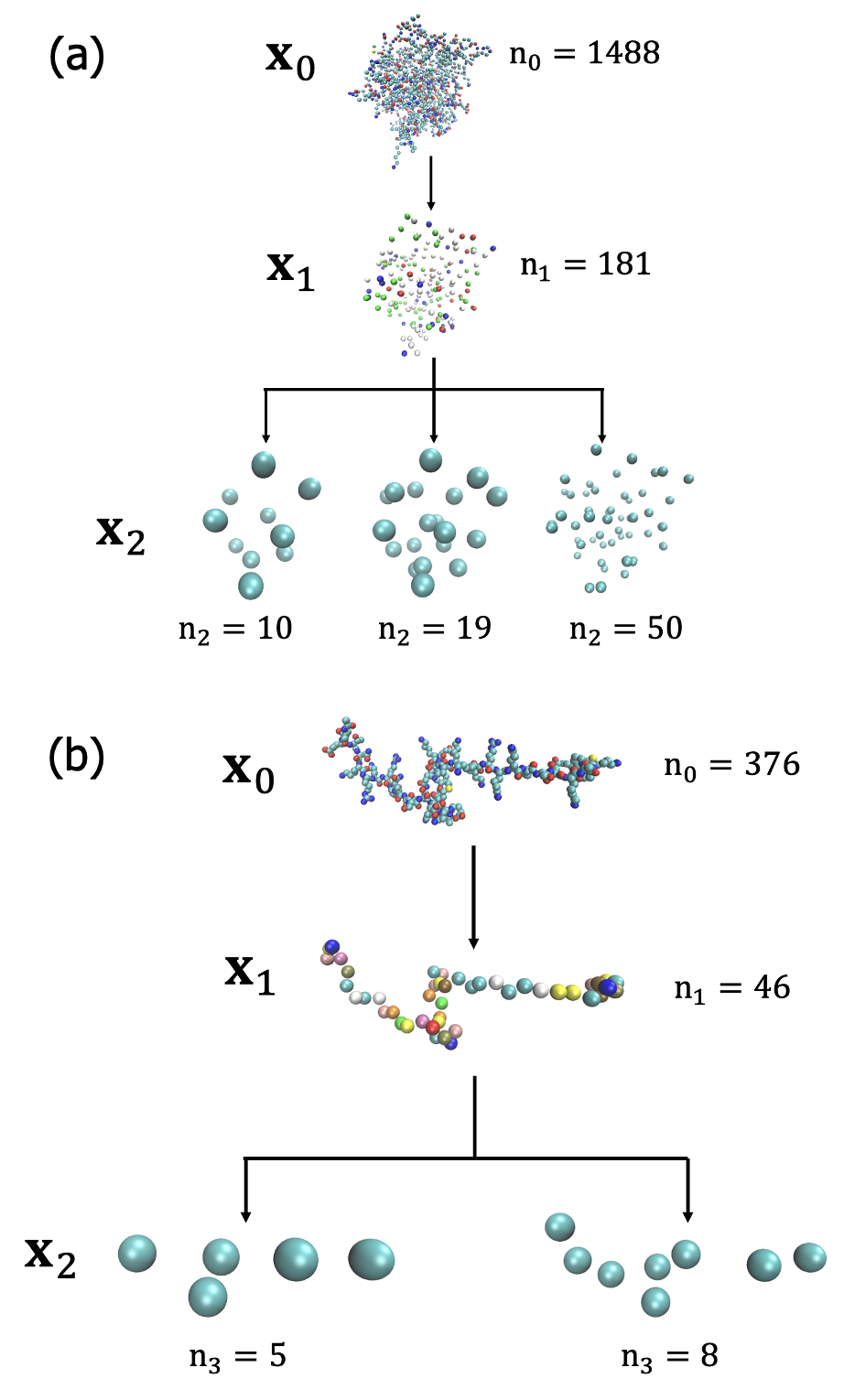}
    \caption{The different FG and CG representations of (a) eIF4E and (b) PED00151 used in our experiments.}
    \label{fig:rep-chain}
\end{figure}

\end{document}